\documentclass[twoside,leqno,twocolumn]{article}

\usepackage[letterpaper]{geometry}

\usepackage{ltexpprt}
\usepackage{hyperref}

\newtheorem{definition}{Definition}
\usepackage{graphicx}
\usepackage{booktabs}
\usepackage{amssymb}
\usepackage{amsmath}
\usepackage{subcaption}
\usepackage{diagbox}
\usepackage{multirow}
\usepackage{multicol}
\usepackage{algorithm}
\usepackage{algorithmic}

\begin{document}

\title{\Large Towards Similarity-Aware Time-Series Classification}
\author{Daochen Zha\thanks{Department of Computer Science, Rice University. \{daochen.zha,khlai,Kaixiong.Zhou,Xia.Hu\}@rice.edu}
\and Kwei-Herng Lai\footnotemark[1] \and Kaixiong Zhou\footnotemark[1] \and Xia Hu\footnotemark[1]}

\date{}

\maketitle


\fancyfoot[R]{\scriptsize{Copyright \textcopyright\ 2022 by SIAM\\
Unauthorized reproduction of this article is prohibited}}





\begin{abstract}
We study time-series classification (TSC), a fundamental task of time-series data mining. Prior work has approached TSC from two major directions: (1) similarity-based methods that classify time-series based on the nearest neighbors, and (2) deep learning models that directly learn the representations for classification in a data-driven manner. Motivated by the different working mechanisms within these two research lines, we aim to connect them in such a way as to jointly model time-series similarities and learn the representations. This is a challenging task because it is unclear how we should efficiently leverage similarity information. To tackle the challenge, we propose \textbf{Sim}ilarity-Aware \textbf{T}ime-\textbf{S}eries \textbf{C}lassification (SimTSC), a conceptually simple and general framework that models similarity information with graph neural networks (GNNs). Specifically, we formulate TSC as a node classification problem in graphs, where the nodes correspond to time-series, and the links correspond to pair-wise similarities. We further design a graph construction strategy and a batch training algorithm with negative sampling to improve training efficiency. We instantiate SimTSC with ResNet as the backbone and Dynamic Time Warping (DTW) as the similarity measure. Extensive experiments on the full UCR datasets and several multivariate datasets demonstrate the effectiveness of incorporating similarity information into deep learning models in both supervised and semi-supervised settings. Our code is 
available at \url{https://github.com/daochenzha/SimTSC}.
\end{abstract}

\maketitle

\section{Introduction}

Time-series classification (TSC) is a fundamental task of time-series data mining. Given a collection of time-series with the attached labels, TSC aims to train a classifier to classify unseen time-series. With the increasing amount of temporal data available, TSC has broad applications, such as human activity recognition, health care, and cyber security~\cite{strodthoff2019detecting}.

Many research efforts have been devoted to TSC. Similarity-based (distance-based) methods are widely used~\cite{abanda2019review}. The main idea is to combine a $k$-NN classifier with a similarity measure for classification. Dynamic Time Warping~(DTW)~\cite{ney1999dynamic,rakthanmanon2012searching}, which calculates the optimal match between two time-series, is one of the most popular similarity measures. It is shown that DTW plus a $1$-NN classifier can achieve reasonably good accuracy~\cite{serra2014empirical}. However, the similarity is often obtained in an unsupervised fashion followed by a simple $k$-NN classifier, which could be sub-optimal. Another promising research line is deep learning. Without any crafting in feature engineering, deep learning methods perform end-to-end training on the raw time-series and learn the representations. Recent studies suggest that convolutional layers, such as ResNet and Fully Convolutional Networks~(FCN), significantly outperform DTW and achieve competitive performance to the state-of-the-art TSC algorithms on the UCR benchmarks~\cite{wang2017time,fawaz2019deep}.

\begin{figure}[t]
  \centering
    \includegraphics[width=0.37\textwidth]{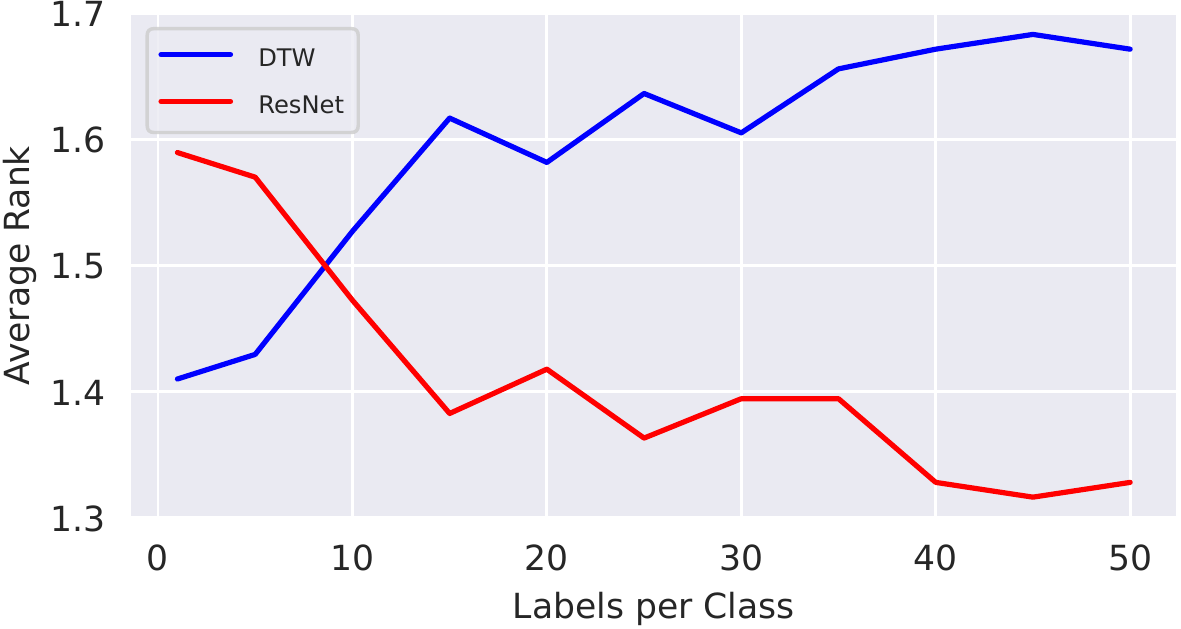}
  \vspace{-10pt}
  \caption{Average ranks ($\downarrow$) of ResNet and DTW on the full 128 UCR datasets, where different numbers of labels per class is given (see Section~\ref{sec:expsetup} for more details).}
  \label{fig:resnetknn}
  \vspace{-10pt}
\end{figure}

While deep learning methods are simple and effective, they highly rely on the supervision of training labels for automatic representation learning; they thus often fall short when very few labels are given. Figure~\ref{fig:resnetknn} compares the average ranks of ResNet and DTW plus $1$-NN on the full 128 UCR datasets~\cite{dau2019ucr} with different numbers of provided training labels per class. While ResNet dominates DTW with sufficient training labels, it delivers unsatisfactory performance with very few labels. In contrast, DTW classifies the time-series by reasoning with pair-wise similarities instead of directly learning time-series representations, which could be less sensitive to the number of provided labels. Motivated by the different working mechanisms within these two research lines, we explore the possibility of connecting them in such a way as to jointly model time-series similarities and learn the representations.


However, it is non-trivial to achieve this goal due to the following challenges. First, it is unclear how we can incorporate similarity information into representation learning. The commonly used architectures, such as CNN and LSTM, cannot model similarity. While some papers have explored neural networks for time-series similarity learning~\cite{abid2018learning,grabocka2018neuralwarp}, they learn the similarity in the first place and then simply apply the learned similarity to classification, which still relies on $k$-NN. Second, even though we can enable similarity in deep learning models, it is challenging to balance similarity information and the original representation learning. Incorporating too much similarity information may lead to indistinguishable representations with poor accuracy.


To address these challenges, we propose \textbf{Sim}ilarity-Aware \textbf{T}ime-\textbf{S}eries \textbf{C}lassification~(SimTSC), a conceptually simple and general framework for incorporating similarity information into deep learning models. Motivated by graph neural networks~(GNNs) in modeling node relationships~\cite{defferrard2016convolutional,hamilton2017inductive,kipf2017semi}, we reformulate TSC as a node classification problem. Specifically, we correspond each time-series to a node in a graph and each pair-wise similarity to a link between nodes. A graph convolution layer is then applied on the top of a backbone (e.g., ResNet) to jointly perform feature extraction and model time-series similarities. We instantiate SimTSC with ResNet and DTW, which are the representative deep learning model and similarity measure, respectively. Extensive experiments suggest that this simple design improves accuracy, particularly with very few labels. We make the following contributions:

\begin{itemize}
    \item Explore the possibility of connecting the research efforts of similarity-based methods and deep learning models for TSC.
    \item Propose SimTSC, a simple and general framework that can combine any similarity measures with any deep learning models from the view of graphs.
    \item Design a graph construction strategy that focuses on the top neighbors for efficient aggregation of graph convolution. We also develop a batch training algorithm with negative sampling to enable the training of SimTSC on large datasets.
    \item Instantiate SimTSC with ResNet and DTW. We conduct extensive experiments on the full 128 UCR datasets and several multivariate datasets. SimTSC outperforms ResNet, DTW, and the state-of-the-art supervised and semi-supervised deep models significantly, particularly when very few labels are given. We also present extensive hyperparameter studies and ablations.
\end{itemize}

\vspace{-5pt}
\section{Preliminaries}

We start with a problem description and then provide a background of time-series similarity measure, deep learning for TSC, and graph neural networks.

\vspace{-5pt}
\subsection{Problem Statement}
\label{sec:21}
We use lowercase alphabet, e.g., $x \in \mathbb{R}$, to represent a scalar value, lowercase boldface letter, e.g., $\mathbf{x}=[x_1, x_2, ..., x_T] \in \mathbb{R}^T$, to denote a vector of length $T$, uppercase boldface alphabet, e.g., $\mathbf{X}=[\mathbf{x}_1, \mathbf{x}_2, ..., \mathbf{x}_M] \in \mathbb{R}^{M \times T}$, to denote a matrix consisting of $M$ vectors, where each vector can have a different length, and calligraphic font, e.g., $\mathcal{X}=[\mathbf{X}_1, \mathbf{X}_2, ..., \mathbf{X}_N] \in \mathbb{R}^{N \times M \times T}$, to denote a 3D matrix. We summarize the main symbols in Table~\ref{tab:symbols}. We first give formal definitions of time-series.
\vspace{-5pt}
\begin{definition}[Univariate Time-Series]
\label{def:1}
A univariate time-series $\mathbf{x}$ of length $T$ is represented as a vector $[x_1, x_2, ..., x_T]$.
\end{definition}
\vspace{-10pt}
\begin{definition}[Multivariate Time-Series]
\label{def:1}
An $M$-dimensional time-series $\mathbf{X}$ consists of $M$ univariate time-series $[\mathbf{x}_1, \mathbf{x}_2, ..., \mathbf{x}_M]$.
\end{definition}
\vspace{-5pt}
Without loss of generality, we unify the above definitions by expanding dimension for univariate time-series. Specifically, we regard a univariate time-series $\mathbf{x}$ as a $1$-dimensional time-series $\mathbf{X} \in \mathbb{R}^{1 \times T}$, that is, a univariate time-series is a special case of multivariate time-series with $M=1$. We will use the unified notation $\mathbf{X}$ to represent a time-series throughout the paper.

\begin{table}[]
    \centering
    \caption{Main symbols and definitions.}
    \vspace{-10pt}
    \scriptsize
    \setlength{\tabcolsep}{0pt}
    \begin{tabular}{l|l} \toprule
    \textbf{Symbol} & \textbf{Definition} \\
    \midrule
    $x \in \mathbb{R}$ & A real-value in a time-series \\
    $\mathbf{x} \in \mathbb{R}^T$ & A univariate time-series with length $T$ \\
    $\mathbf{X} \in \mathbb{R}^{M \times T}$ & An $M$-dimensional time-series with length $T$ \\
    $\mathcal{X} \in \mathbb{R}^{N \times M \times T}$ & A 3D matrix consisting of $N$ multivariate time-series\\
    $\mathcal{X}^\text{train}$ & A set of training time-series \\
    $\mathbf{y}^\text{train}$ & The labels corresponding to $\mathcal{X}^\text{train}$ \\
    $\mathcal{X}^\text{test}$ & A set of testing time-series \\
    $\mathbf{y}^\text{test}$ & The labels corresponding to $\mathcal{X}^\text{test}$ \\
    $\mathcal{X}^\text{unlabeled}$ & A set of unlabeled time-series \\
    $d(\mathbf{X}_1, \mathbf{X}_2)$ & The similarity (distance) of two time-series \\
    $\mathbf{D} \in \mathbb{R}^{N \times N}$ & The similarity (distance) matrix \\
    $\widetilde{\mathbf{A}} \in \mathbb{R}^{N \times N}$ & The normalized adjacency matrix in graph \\
    $\widetilde{\mathbf{X}} \in \mathbb{R}^{N \times M}$ & The attribute information matrix in graph \\
    $\alpha$ & A scaling factor \\
    $K$ & The number of neighbors for each node \\

    \bottomrule
    \end{tabular}
    \label{tab:symbols}
    \vspace{-15pt}
\end{table}

We formally describe the problem of TSC. Given some testing time-series $\mathcal{X}^\text{test}=[\mathbf{X}_1, \mathbf{X}_2, ..., \mathbf{X}_{N^\text{test}}]$ and the labels $\mathbf{y}^\text{test}=[y_1, y_2, ..., y_{N^\text{test}}]$, where $N^\text{test}$ is the number of testing time-series, we aim to train a classifier that can predict the labels based on $\mathcal{X}^\text{test}$ under one of the following settings:
\vspace{-5pt}
\begin{itemize}
    \item \textbf{Supervised setting}: The classifier is trained based on a training time-series dataset $\mathcal{X}^\text{train}=[\mathbf{X}_1, \mathbf{X}_2, ..., \mathbf{X}_{N^\text{train}}]$ and its corresponding labels $\mathbf{y}^\text{train}=[y_1, y_2, ..., y_{N^\text{train}}]$, where $N^\text{train}$ is the number of training time-series.
    \item \textbf{Inductive semi-supervised setting}: In addition to $\mathcal{X}^\text{train}$ and $\mathbf{y}^\text{train}$, the classifier can also access some unlabeled time-series  $\mathcal{X}^\text{unlabeled}$, which does not overlap with $\mathcal{X}^\text{test}$.
    \item \textbf{Transductive semi-supervised setting}: In addition to $\mathcal{X}^\text{train}$, $\mathbf{y}^\text{train}$ and $\mathcal{X}^\text{unlabeled}$, the classifier is exposed to testing time-series $\mathcal{X}^\text{test}$. Note that $\mathbf{y}^\text{test}$ is not accessible in training.
\end{itemize}
\vspace{-5pt}
The above settings differ in how much unlabeled time-series data the classifier can access. With more unlabeled data, the classifier could better learn the underlying data distributions and often achieve better accuracy.

\vspace{-5pt}
\subsection{Time-Series Similarity Measure}
Given two time-series $\mathbf{X}_1$ and $\mathbf{X}_2$, we aim at providing a distance $d(\mathbf{X}_1, \mathbf{X}_2)$, such that similar time-series tend to have smaller $d(\mathbf{X}_1, \mathbf{X}_2)$. Dynamic Time Warping~(DTW)~\cite{ney1999dynamic} is one of the most popular ones. The key idea is to calculate the optimal match between two time-series such that the sum of matched series has the smallest values. The troughs and peaks of the same pattern can be matched even if they are not perfectly synced up. DTW is a standard tool with many efficient implementations, such as the UCR Suite~\cite{rakthanmanon2012searching}. In this work, we adopt DTW as the similarity measure; one can also use other similarity measures under our framework.

\vspace{-5pt}
\subsection{Deep Neural Networks for TSC}
Numerous deep learning models have been developed for time-series classification~\cite{wang2017time,karim2017lstm,karim2019multivariate,fawaz2020inceptiontime}. In this work, we mainly focus on Residual Network~(ResNet) since it is shown to have superior performance on the majority of UCR Time Series Classification Archive~\cite{fawaz2019deep}. The network consists of multiple residual blocks. Each block consists of three 1D convolutional layers followed by batch normalization and a ReLU activation function, with shortcuts to enable a direct flow of the gradient:
\begin{align}
\vspace{-3pt}
 & \mathbf{H}_1 = \text{ReLU}(\text{BatchNorm}(\text{Conv1d}(\mathbf{X}))), \label{eqn:1}\\
 & \mathbf{H}_2 = \text{ReLU}(\text{BatchNorm}(\text{Conv1d}(\mathbf{H}_1))), \label{eqn:2}\\
 & \mathbf{H}_3 = \text{ReLU}(\text{BatchNorm}(\text{Conv1d}(\mathbf{H}_2))), \label{eqn:3}\\
 & \hat{\mathbf{H}} = \text{ReLU}(\mathbf{H}_3 + \mathbf{X}), \label{eqn:4}
\vspace{-3pt}
\end{align}
where $\hat{\mathbf{H}}$ is the output of residual block, $\text{Conv1d}(\cdot)$ denotes 1D convolutional, $\text{BatchNorm}(\cdot)$ is batch normalization, and $\text{ReLU}(\cdot)$ is ReLU activation function.

\subsection{Graph Neural Networks}
Graph neural networks~(GNNs) have achieved great success in modeling node dependencies in graph~\cite{defferrard2016convolutional,kipf2017semi,zhou2020multi}. To capture the node dependency, Graph Convolution Network~(GCN)~\cite{kipf2017semi} performs joint learning of feature extraction and aggregation of neighboring nodes. Given a graph $G=(\widetilde{\mathbf{A}}, \widetilde{\mathbf{X}})$, where $\widetilde{\mathbf{A}} \in \mathbb{R}^{N \times N}$ is the normalized adjacency matrix, $\widetilde{\mathbf{X}} \in \mathbb{R}^{N \times M}$ is the attribute information matrix, $N$ is the number of nodes, and $M$ is the feature dimension, a GCN layer performs feature aggregation of neighboring nodes with
\begin{equation}
    \label{eq:5}
    \hat{\mathbf{Z}} = \widetilde{\mathbf{A}}\widetilde{\mathbf{X}}\mathbf{W},
\end{equation}
where $\mathbf{W}$ denotes trainable parameters, and $\hat{\mathbf{Z}}$ denotes the output of the GCN layer. Similar to ResNet, we can stack multiple GCN layers with activation functions. While there are hundreds of GNNs, we adopt the basic GCN to make our contribution focused; one can also use other GNNs under our framework.

\section{Methodology}
\begin{figure*}[t]
  \centering
    \includegraphics[width=0.7\textwidth]{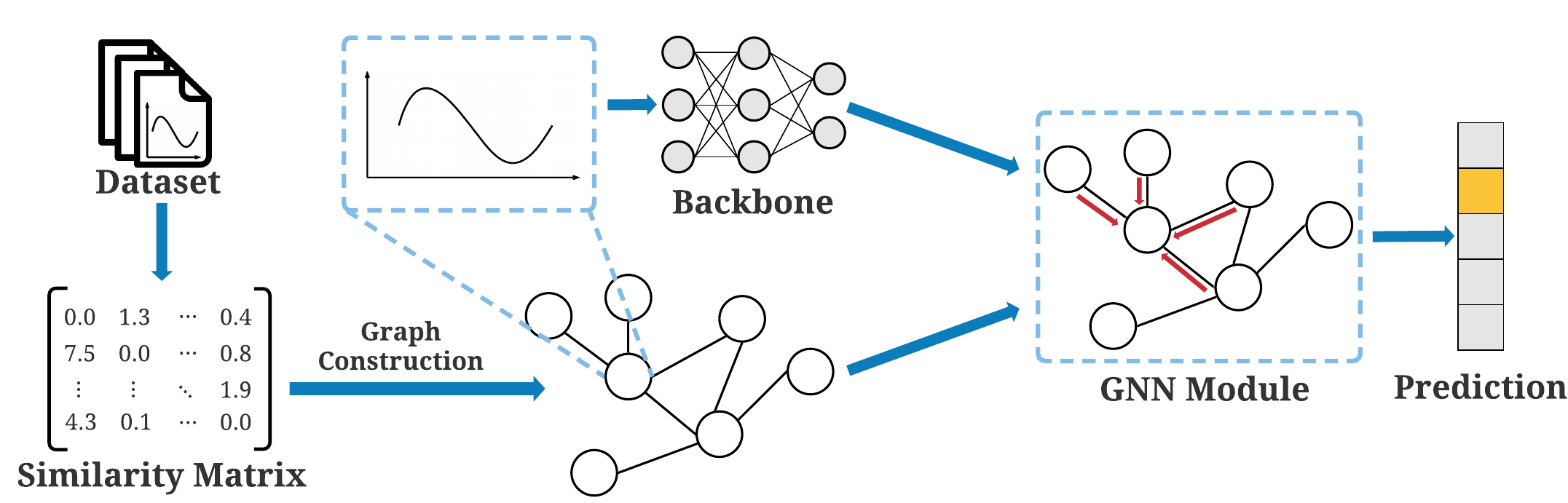}
  \vspace{-10pt}
  \caption{An overview of SimTSC framework. The graph is constructed based on the pair-wise similarities (e.g., DTW distances) of the time-series. Each time-series is processed by a backbone (e.g., ResNet) for feature extraction. The GNN module will aggregate the features and produce the final representations for classification.}
  \label{fig:overview}
  \vspace{-10pt}
\end{figure*}

Figure~\ref{fig:overview} shows an overview of \textbf{Sim}ilarity-Aware \textbf{T}ime-\textbf{S}eries \textbf{C}lassification (SimTSC), which consists of three modules: (1) a graph construction module that connects the time-series based on a similarity measure (e.g., DTW), (2) a backbone that performs feature extraction with deep neural networks (e.g., ResNet), and (3) a GNN module that aggregates the features of neighboring time-series (e.g., GCN). The graph construction is unsupervised so that it can flexibly adapt to all the three settings defined in Section~\ref{sec:21}.


\subsection{Graph Construction with Similarity}
This subsection describes how we construct a graph based on a similarity measure of time-series, e.g., DTW.


Let $\mathcal{X} = [\mathbf{X}_1, \mathbf{X}_2, ..., \mathbf{X}_N]$ denote all the accessible time-series. In the supervised setting, $\mathcal{X}$ is simply all the training data. In the semi-supervised settings defined in Section~\ref{sec:21}, the $\mathcal{X}$ consists of both labeled and unlabeled time-series. Based on the learned similarity measure $d(\cdot, \cdot)$, we can obtain a similarity matrix for $\mathcal{X}$ as
\begin{equation}
    \label{eqn:6}
    \mathbf{D} = \begin{bmatrix} d(\mathbf{X}_1,\mathbf{X}_1)  & d(\mathbf{X}_1,\mathbf{X}_2) & \cdots & d(\mathbf{X}_1,\mathbf{X}_N)\\
                                 d(\mathbf{X}_2,\mathbf{X}_1)  & d(\mathbf{X}_2,\mathbf{X}_2) & \cdots & d(\mathbf{X}_2,\mathbf{X}_N)\\
                                 \vdots  & \vdots & \ddots & \vdots\\
                                 d(\mathbf{X}_N,\mathbf{X}_1)  & d(\mathbf{X}_N,\mathbf{X}_2) & \cdots & d(\mathbf{X}_N,\mathbf{X}_N)\\
                  \end{bmatrix}.
\end{equation}

Given the similarity matrix $\mathbf{D}$, we construct the graph as follows. First, we introduce a scaling hyperparameter $\alpha \in [0, \infty)$ to control the importance of top neighbors. Specifically, let $\mathbf{D}_{ij}$ denote the $(i,j)^{th}$ entry of $\mathbf{D}$. The adjacency matrix $\mathbf{A}$ is obtained by
\begin{equation}
\label{eqn:7}
    \mathbf{A}_{ij} = \frac{1}{e^{\alpha \mathbf{D}_{ij}}}, \forall i, j,
\end{equation}
where $\mathbf{A}_{ij}$ denotes the $(i,j)^{th}$ entry of $\mathbf{A}$. A larger $\alpha$ will give more importance to the top neighbors. When $\alpha=0$, each node will equally aggregate the features of all the nodes, and all the nodes will have indistinguishable features. When $\alpha \to \infty$, our framework reduces to the backbone itself since $\mathbf{A}$ reduces to a diagonal matrix.

Second, to filter out irrelevant neighbors, we sample the top-$K$ neighbors for each node. Specifically, for each row $\mathbf{a}_i$ in $\mathbf{A}$, we only keep $K$ entries with the largest weights and zero out the others, which leads to a sparse matrix. Finally, we normalize the adjacency matrix with $\widetilde{\mathbf{A}}_{ij} = \frac{\mathbf{A}_{ij}}{\sum_{j'} \mathbf{A}_{ij'}}, \forall i,j$.

\subsection{Joint Learning of Backbone and Graph Convolution Layers}
This subsection introduces how we optimize the backbone and the GNN module.

The backbone takes all the accessible time-series $\mathcal{X}$ as inputs and produces an attribute information matrix $\widetilde{\mathbf{X}} \in \mathbb{R}^{N \times M}$, where $N$ denotes the number of all the accessible time-series, and $M$ denotes the feature dimension. Here, the output of the backbone is assumed to be flattened to be 1-dimensional. Given $\widetilde{\mathbf{X}}$ and $\widetilde{\mathbf{A}}$, the GNN aggregates the node features and produces the final representation followed by a softmax layer for classification.

Let $\widetilde{\mathbf{Z}} \in \mathbb{R}^{N \times C}$ denote the final output of the model, where $C$ is the number classes, and $\widetilde{\mathbf{Z}}^\text{train} \in \mathbb{R}^{N^\text{train} \times C}$ denotes the rows that have labels. The objective is to minimize the cross-entropy over the labeled time-series:
\begin{equation}
\label{eqn:8}
    \mathcal{L} = - \sum_{i=1}^{N^\text{train}} \sum_{j=1}^{C} \mathbf{Y}_{ij}^\text{train} \log \widetilde{\mathbf{Z}}_{ij}^\text{train},
\end{equation}
where $\mathbf{Y}_{ij}^\text{train}$ denotes the $j^{th}$ element of one-hot encoded label of the $i^{th}$ labeled time-series. The weights of the backbone and the GNN module can be jointly optimized using gradient descent.

\subsection{Batch Training with Negative Sampling}
This subsection presents how we handle large datasets with batch training and how we use negative sampling to improve efficiency.

A naive training strategy is to put all the time-series into the GPU memory in the first place and then train the network with backpropagation. However, this strategy has the following limitations. First, it cannot scale to large datasets with many time-series or very long time-series. Second, each time-series can only interact with a fixed number of neighbors since the constructed graph is static. However, the top-$K$ dropping strategy may improperly drop some important connections, which leads to sub-optimal performance.

To overcome the above limitations, we propose a batch training strategy with negative sampling as follows. Given a batch size $B$, we sample $B/2$ labeled and $B/2$ unlabeled time-series, respectively, as a batch for training. The unlabeled time-series are ``negatively sampled" in that we can usually access far more unlabeled data in real-world applications. Then, we construct a graph with this batch of time-series and update the model accordingly. The above training procedure is summarized in Algorithm~\ref{alg:1}. Here, $\mathcal{X}^\text{unlabeled}$ will also cover testing data in the transductive setting\footnote{In the supervised setting, we only sample labeled time-series.}.

This design has three nice properties. First, it can scale to large datasets since we only need to put a batch of data into GPU memory in each update step. Second, each node will interact with more neighbors for aggregation since a new graph will be constructed in every randomly sampled batch. Third, compared with random sampling, negative sampling ensures that half of the data in each batch is labeled so that we can have sufficient learning signals to update the weights. In particular, if very few labels are available, random sampling may result in very few or even no labeled data in a sampled batch, which will reduce training efficiency.

At testing time, we similarly sample $B/2$ testing and $B/2$ non-testing time-series in each batch to construct the graph. In this way, we can aggregate some of the non-testing representations to enhance the representations of the testing samples.

\vspace{-5pt}
\section{Experiments}

We evaluate SimTSC across various settings to answer the following research questions: \textbf{RQ1}: How does SimTSC compare with existing deep learning and similarity-based methods (Section~\ref{sec:exp1})? \textbf{RQ2:} Can SimTSC also enhance other neural architectures, such as MLP and FCN (Section~\ref{sec:exp2})? \textbf{RQ3:} How will the number of graph convolutional layers impact the performance~(Section~\ref{sec:exp3})? \textbf{RQ4:} How will graph structure impact the performance of SimTSC~(Section~\ref{sec:exp4})? \textbf{RQ5:} Is the proposed negative sampling strategy effective~(Section~\ref{sec:exp5})? \textbf{RQ6:} Can SimTSC also be applied to multivariate time-series classification tasks (Section~\ref{sec:exp6})? \textbf{RQ7:} How does SimTSC learn the representations with similarity information~(Section~\ref{sec:exp7})?

\begin{algorithm}[t]
\caption{Batch training of SimTSC with negative sampling}
\label{alg:1}
\setlength{\intextsep}{0pt} 
\begin{algorithmic}[1]
\STATE \textbf{Input:} Labeled time-series $\mathcal{X}^\text{train}$, unlabeled time-series $\mathcal{X}^\text{unlabeled}$, similarity measure $d(\cdot, \cdot)$, batch size $B$
\STATE $\mathcal{X} \leftarrow \{\mathcal{X}^\text{train}, \mathcal{X}^\text{unlabeled}\}$
\STATE Precompute the similarity matrix $\mathbf{D}$ for $\mathcal{X}$ based on Eq.~\ref{eqn:6}
\FOR{each epoch}
    \FOR{each batch $\mathcal{X}^\text{train}_\text{batch}$ of size $B/2$ in $\mathcal{X}^\text{train}$}
        \STATE Sample a batch $\mathcal{X}^\text{unlabeled}_\text{batch}$ of size $B/2$ from $\mathcal{X}^\text{unlabeled}$
        \STATE $\mathcal{X}_\text{batch} \leftarrow \{\mathcal{X}^\text{train}_\text{batch}, \mathcal{X}^\text{unlabeled}_\text{batch}\}$
        \STATE Obtain submatrix $\mathbf{D}_\text{batch}$ from $\mathbf{D}$ with sampled indices
        \STATE Construct adjacency matrix $\mathbf{A}$ from $\mathbf{D}_\text{batch}$ based on Eq.~\ref{eqn:7}
        \STATE Compute normalized matrix $\widetilde{\mathbf{A}}$ with $\widetilde{\mathbf{A}}_{ij} = \frac{\mathbf{A}_{ij}}{\sum_{j'} \mathbf{A}_{ij'}}, \forall i,j$
        \STATE Obtain model output $\widetilde{\mathbf{Z}}$ with $\mathcal{X}_\text{batch}$ and $\widetilde{\mathbf{A}}$
        \STATE Update the model weights based on Eq.~\ref{eqn:8}
    \ENDFOR
\ENDFOR
\end{algorithmic}
\end{algorithm}

\vspace{-5pt}
\subsection{Experimental Setup}
\label{sec:expsetup}
As suggested in~\cite{dau2019ucr}, we evaluate the performance on the full 128 UCR datasets\footnote{\url{https://www.cs.ucr.edu/~eamonn/time_series_data_2018/}}. We merge the original training and testing data to create new splits for all the datasets to simulate the three settings defined in Section~\ref{sec:21}. First, we randomly split 20\% of the data as the hold-out set for testing purposes, denoted as $\mathcal{X}^\text{test}$. Second, we vary the number of training labels per class from the set $\{1, 5, 10, 15, 20, 25, 30, 35, 40, 45, 50\}$ to test different levels of supervision and sample a subset from the remaining 80\% of data as the training data to create few-shot settings, denoted as $\mathcal{X}^\text{train}$. Third, the time-series out of the above two splits will serve as another split of some unlabeled data, denoted as $\mathcal{X}^\text{unlabeled}$. We further consider four multivariate datasets: Character Trajectories~\cite{bagnall2018uea}, ECG~\cite{olszewski2001generalized}, KickVsPunch~\cite{baydogan2019multivariate}, and NetFlow~\cite{baydogan2019multivariate}.

\textbf{Metric.} For UCR datasets, we follow previous work~\cite{dau2019ucr,fawaz2019deep} and rank the algorithms on each dataset according to the mean accuracy and report the average ranks across the 128 datasets. We perform Wilcoxon signed-rank test with a significance level of $0.05$. For the multivariate datasets, we report accuracies.

\textbf{Baselines.} For deep learning models, we consider four \textbf{supervised} architectures, including MLP, Fully Convolutional Network (FCN)~\cite{wang2017time}, ResNet~\cite{geng2018cost}, and InceptionTime~\cite{fawaz2020inceptiontime}, and a state-of-the-art \textbf{semi-supervised} framework TapNet~\cite{zhang2020tapnet}. For \textbf{similarity-based} methods, we include DTW with a $1$ Nearest-Neighbor ($1$-NN) classifier~\cite{serra2014empirical}. We train SimTSC in supervised, inductive semi-supervised, and transductive semi-supervised settings, denoted as SimTSC-S, SimTSC-I, and SimTSC-T, respectively.

\begin{table*}[t]
    \centering
    \scriptsize
    \caption{Average ranks ($\downarrow$) of SimTSC and baselines on the full 128 UCR datasets with different numbers of training labels per class. $\dagger$, $\blacktriangle$, and $\triangledown$ to denote the cases where SimTSC-S, SimTSC-I, and SimTSC-T are significantly better than the other algorithms w.r.t. the Wilcoxon signed rank test ($p < 0.05$), respectively.}
    \label{tab:performance}
    \vspace{-10pt}
    \setlength{\tabcolsep}{3pt}
    \begin{tabular}{l|lllllllllll}
    \toprule
     
 \diagbox [width=7em,trim=l] {Algorithm}{Labels}& 1 & 5 & 10 & 15 & 20 & 25 & 30 & 35 & 40 & 45 & 50 \\
    \midrule
    \midrule
     DTW & 3.776 & 4.163 & 4.465$\blacktriangle$$\triangledown$ & 4.738$\blacktriangle$$\triangledown$ & 4.824$\dagger$$\blacktriangle$$\triangledown$ & 5.048$\dagger$$\blacktriangle$$\triangledown$ & 4.965$\dagger$$\blacktriangle$$\triangledown$ & 5.160$\dagger$$\blacktriangle$$\triangledown$ & 5.309$\dagger$$\blacktriangle$$\triangledown$ & 5.199$\dagger$$\blacktriangle$$\triangledown$ & 5.211$\dagger$$\blacktriangle$$\triangledown$ \\
     MLP & 5.504$\dagger$$\blacktriangle$$\triangledown$ & 5.496$\dagger$$\blacktriangle$$\triangledown$ & 5.438$\dagger$$\blacktriangle$$\triangledown$ & 5.309$\dagger$$\blacktriangle$$\triangledown$ & 5.316$\dagger$$\blacktriangle$$\triangledown$ & 5.256$\dagger$$\blacktriangle$$\triangledown$ & 5.367$\dagger$$\blacktriangle$$\triangledown$ & 5.477$\dagger$$\blacktriangle$$\triangledown$ & 5.195$\dagger$$\blacktriangle$$\triangledown$ & 5.402$\dagger$$\blacktriangle$$\triangledown$ & 5.348$\dagger$$\blacktriangle$$\triangledown$ \\
     FCN & 4.630$\blacktriangle$$\triangledown$ & 4.310 & 4.383$\blacktriangle$$\triangledown$ & 4.508$\blacktriangle$$\triangledown$ & 4.723$\dagger$$\blacktriangle$$\triangledown$ & 4.803$\dagger$$\blacktriangle$$\triangledown$ & 4.699$\dagger$$\blacktriangle$$\triangledown$ & 4.910$\dagger$$\blacktriangle$$\triangledown$ & 4.773$\dagger$$\blacktriangle$$\triangledown$ & 4.883$\dagger$$\blacktriangle$$\triangledown$ & 4.852$\dagger$$\blacktriangle$$\triangledown$ \\
     ResNet & 4.846$\dagger$$\blacktriangle$$\triangledown$ & 4.857$\dagger$$\blacktriangle$$\triangledown$ & 4.617$\dagger$$\blacktriangle$$\triangledown$ & 4.047 & 4.449$\blacktriangle$$\triangledown$ & 4.039 & 4.102 & 4.090 & 4.086 & 3.840 & 3.895 \\
     InceptionTime & 5.484$\dagger$$\blacktriangle$$\triangledown$ & 5.302$\dagger$$\blacktriangle$$\triangledown$ & 5.438$\dagger$$\blacktriangle$$\triangledown$ & 5.434$\dagger$$\blacktriangle$$\triangledown$ & 5.215$\dagger$$\blacktriangle$$\triangledown$ & 5.145$\dagger$$\blacktriangle$$\triangledown$ & 5.168$\dagger$$\blacktriangle$$\triangledown$ & 4.914$\dagger$$\blacktriangle$$\triangledown$ & 4.941$\dagger$$\blacktriangle$$\triangledown$ & 5.066$\dagger$$\blacktriangle$$\triangledown$ & 5.039$\dagger$$\blacktriangle$$\triangledown$ \\
     \midrule
     SimTSC-S & 4.224$\blacktriangle$ & 4.278$\blacktriangle$$\triangledown$ & 4.074 & 4.277$\blacktriangle$$\triangledown$ & 4.141$\blacktriangle$$\triangledown$ & 4.044$\triangledown$ & 4.148$\blacktriangle$ & 3.988 & 3.887 & 3.918 & 4.047 \\
     SimTSC-I & \textbf{3.724} & 3.817 & 3.793 & \textbf{3.836} & 3.746 & 4.031$\blacktriangle$$\triangledown$ & \textbf{3.762} & 3.734 & \textbf{3.852} & 3.867 & \textbf{3.797} \\
     SimTSC-T & 3.811 & \textbf{3.778} & \textbf{3.781} & 3.852 & \textbf{3.586} & \textbf{3.632} & 3.789 & \textbf{3.727} & 3.957 & \textbf{3.824} & 3.812 \\
   
     \bottomrule
    \end{tabular}
    \vspace{-10pt}
\end{table*}

\textbf{Implementation Details.} We use DTW as the similarity measure and ResNet as the backbone. The hyperparamters are set based on the accuracy on the training data, with the scaling factor $\alpha$ as 0.3, the number of neighbors $K$ as 3, one GCN layer, the batch size as 128, and the number of epochs to be 500, across all the experiments. For a fair comparison, the backbones used in SimTSC are exactly the same as the baselines. We use the authors' implementations of InceptionTime\footnote{\url{https://github.com/hfawaz/InceptionTime}} and TapNet\footnote{\url{https://github.com/xuczhang/tapnet}} with the default hyperparameters. We run five times and report the average accuracy. More details of the neural architectures, hardware, and the dataset statistics are provided in supplementary materials.

\begin{table}[t]
    \centering
    \scriptsize
    \caption{Average ranks of TapNet and SimTSC on 83 datasets, on which TapNet does not suffer from memory explosion. $\triangledown$ suggests SimTSC is significantly better.}
    \label{tab:tapnet}
    \vspace{-10pt}
    \setlength{\tabcolsep}{6.0pt}
    \begin{tabular}{l|lllll}
    \toprule
     
 \diagbox [width=8em,trim=l] {Algorithm}{Labels}& 5 & 10 & 15 & 20 & 25 \\
    \midrule
    \midrule
     TapNet & 1.645$\triangledown$ & 1.524 & 1.548 & 1.536 & 1.530\\
     SimTSC & \textbf{1.355} & \textbf{1.476} & \textbf{1.452} & \textbf{1.464} & \textbf{1.470} \\

     \bottomrule
    \end{tabular}
    \vspace{-15pt}
\end{table}

\vspace{-5pt}
\subsection{Performance Comparison on Benchmarks}
\label{sec:exp1}
To study \textbf{RQ1}, we report the average ranks of SimTSC on the UCR datasets in Table~\ref{tab:performance}, separately present TapNet in Table~\ref{tab:tapnet} because TapNet does not support batch training and suffers from memory explosion on 45 datasets. We make the following observations.

\textbf{First}, SimTSC outperforms the baselines with very few labels. Given 1, 5, or 10 labels per class, SimTSC trained in all the settings achieve better ranks than ResNet significantly. Given more labels, i.e, 15, 20, 25, 30 , or 35, SimTSC also beats ResNet consistently. An interesting observation is that SimTSC-S performs well even though it only uses very few time-series to construct the graph. A possible explanation is that the graph may serve as a regularizer for ResNet to help alleviate the overfitting issue. \textbf{Second}, similarity information is less effective when we have sufficient labels. Given 40, 45, or 50 labels, ResNet and SimTSC achieve similar ranks. Nevertheless, the results suggest that SimTSC can still deliver competitive performance with enough labels, which shows the flexibility of SimTSC. \textbf{Third}, SimTSC tends to perform better with more unlabeled data. With very few exceptions, SimTSC-T $>$ SimTSCI $>$ SimTSC-S. This is because we can construct a better graph to capture the underlying data distributions with more unlabeled time-series. \textbf{Fourth}, SimTSC consistently outperforms TapNet across all different numbers of labels when both evaluated under transductive semi-supervised setting. The result again verifies the effectiveness of SimTSC.

\begin{table}[t]
    \centering
    \scriptsize
    \caption{Average ranks of using MLP and FCN as backbones. $\dagger$, $\blacktriangle$, and $\triangledown$ suggest SimTSC-S, SimTSC-I, and SimTSC-T are significantly better, respectively.}
    \label{tab:mlcfcn}
    \vspace{-10pt}
    \setlength{\tabcolsep}{0.0pt}
    \begin{tabular}{l|llll}
    \toprule
     
 \diagbox [width=14em,trim=l] {Algorithm}{Labels}& 10 & 20 & 30 & 40 \\
    \midrule
    \midrule
     MLP & 3.137$\dagger$$\blacktriangle$$\triangledown$ & 2.906$\dagger$$\blacktriangle$$\triangledown$ & 2.879$\dagger$$\blacktriangle$$\triangledown$ & 2.898$\dagger$$\blacktriangle$$\triangledown$ \\
     SimTSC-S with MLP backbone & 2.738$\blacktriangle$$\triangledown$ & 2.633$\blacktriangle$$\triangledown$ & 2.618$\blacktriangle$$\triangledown$ & 2.578$\blacktriangle$$\triangledown$ \\
     SimTSC-I with MLP backbone & 2.078 & 2.258 & 2.414$\triangledown$ & \textbf{2.258} \\
     SimTSC-T with MLP backbone & \textbf{2.047} & \textbf{2.203} & \textbf{2.090} & 2.266 \\
     
     \midrule
     FCN & 2.672 & 2.695$\triangledown$ & 2.641 & 2.563 \\
     SimTSC-S with FCN backbone & 2.523 & 2.492 & 2.504 & \textbf{2.438} \\
     SimTSC-I with FCN backbone & 2.500$\triangledown$ & 2.590$\triangledown$ & 2.539 & 2.547 \\
     SimTSC-T with FCN backbone & \textbf{2.309} & \textbf{2.223} & \textbf{2.316} & 2.453 \\

     \bottomrule
    \end{tabular}
    \vspace{-13pt}
\end{table}

\subsection{Results on Other Neural Architectures}
\label{sec:exp2}
To investigate \textbf{RQ2}, we show the results of applying SimTSC on MLP and FCN on the UCR datasets in Table~\ref{tab:mlcfcn}. SimTSC can also enhance these two neural architectures. In particular, we observe significant performance gains when applying SimTSC on MLP.

\begin{figure}[t]
  \centering
  \begin{subfigure}[b]{0.20\textwidth}
    \centering
    \includegraphics[width=0.99\textwidth]{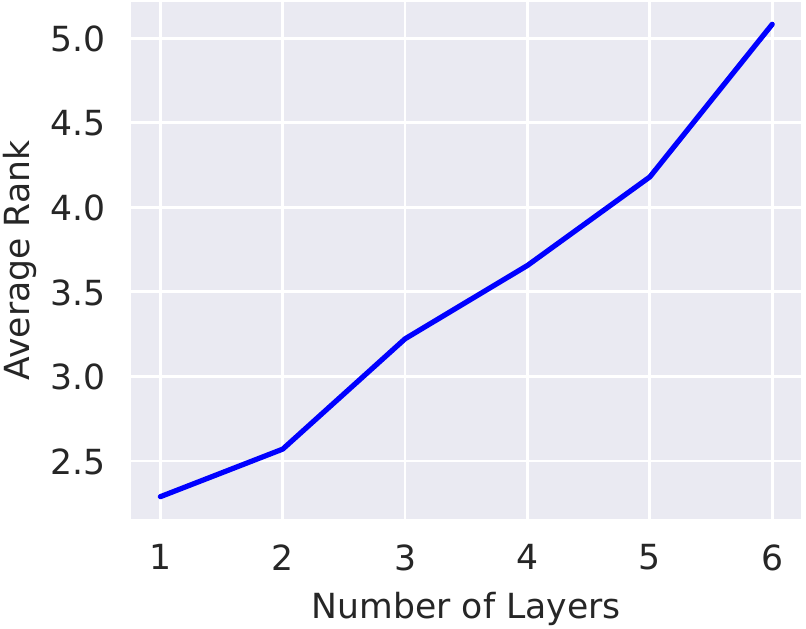}
    \caption{10 labels per class}
    \vspace{-10pt}
  \end{subfigure}%
  \begin{subfigure}[b]{0.20\textwidth}
    \centering
    \includegraphics[width=0.99\textwidth]{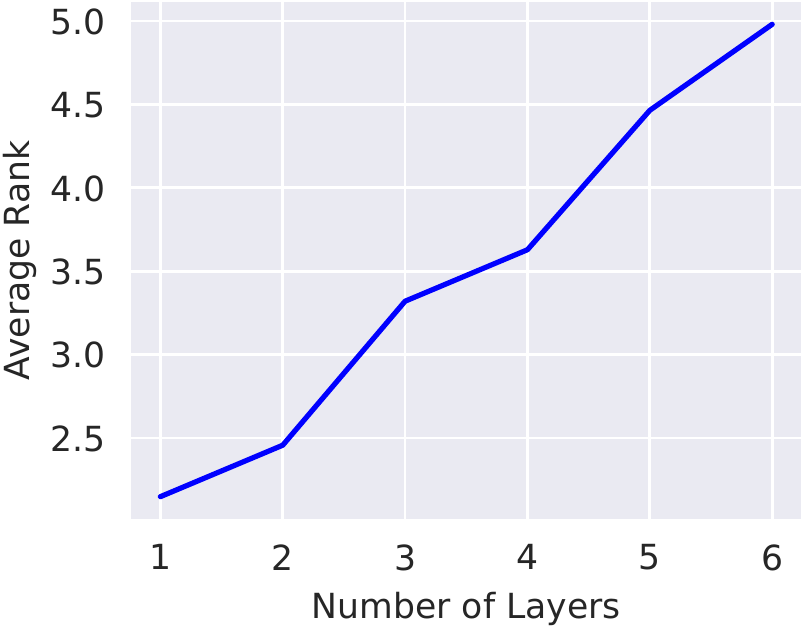}
    \caption{20 labels per class}
    \vspace{-10pt}
  \end{subfigure}%
  \caption{Impact of the number of GCN layers.}
  \label{fig:impactlayer}
  \vspace{-10pt}
\end{figure}

\subsection{Analysis of Graph Convolutional Layers}
\label{sec:exp3}
To answer \textbf{RQ3}, we report the ranks using more GCN layers. We focus on the transductive setting with 10 and 20 labels per class (Figure~\ref{fig:impactlayer}). SimTSC achieves the best performance with only one GCN layer. We speculate that this is because the graph is constructed based on the pair-wise similarity, which can fully capture the relationship between each pair of time-series. As such, stacking more GCN layers will not introduce more information but instead makes the model more susceptible to over-smoothing~\cite{zhou2020towards,zhou2021dirichlet}. Nevertheless, stacking more GCN layers could help in larger time-series datasets by computing a submatrix and leverage multiple GCN layers to capture the multi-hop connections, which is deferred as our future work.

\begin{figure}[t]
  \centering
  \begin{subfigure}[b]{0.20\textwidth}
    \centering
    \includegraphics[width=0.99\textwidth]{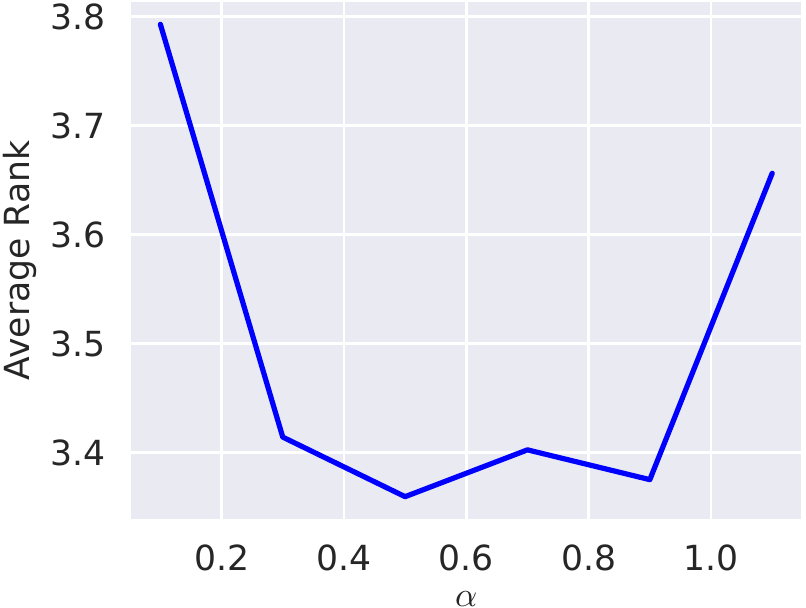}
    \caption{10 labels per class}
  \vspace{-10pt}
  \end{subfigure}%
  \begin{subfigure}[b]{0.20\textwidth}
    \centering
    \includegraphics[width=0.99\textwidth]{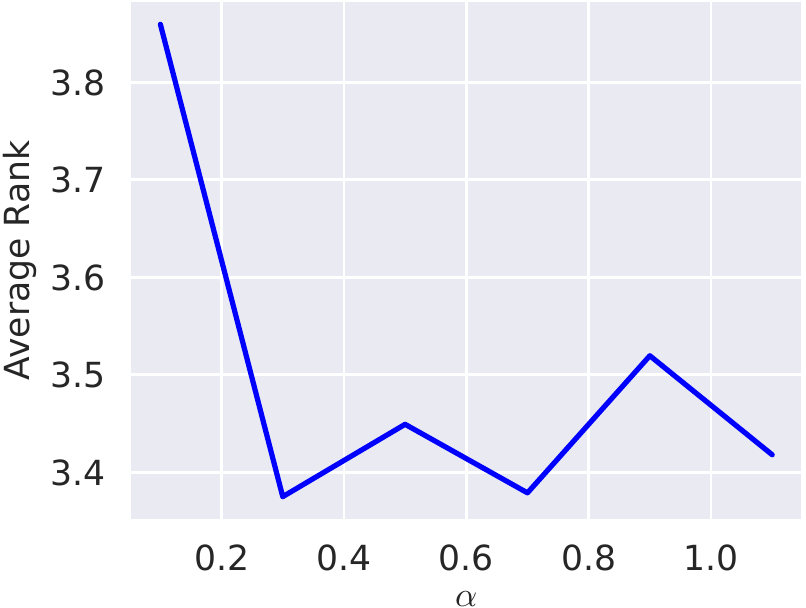}
    \caption{20 labels per class}
  \vspace{-10pt}
  \end{subfigure}%
  \caption{Impact of $\alpha$.}
  \vspace{-8pt}
  \label{fig:impactalpha}
\end{figure}

\begin{figure}[t]
  \centering
  \begin{subfigure}[b]{0.200\textwidth}
    \centering
    \includegraphics[width=0.99\textwidth]{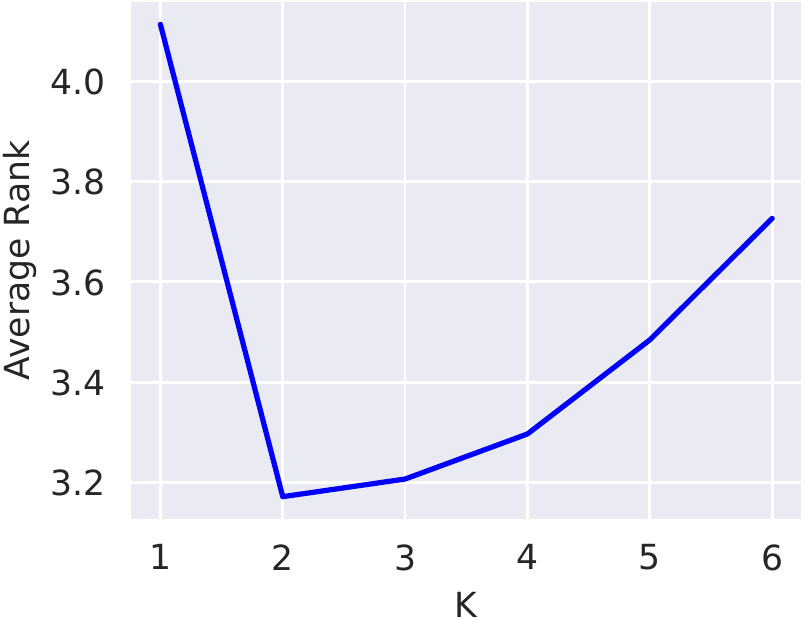}
    \caption{10 labels per class}
    \vspace{-10pt}
  \end{subfigure}%
  \begin{subfigure}[b]{0.20\textwidth}
    \centering
    \includegraphics[width=0.99\textwidth]{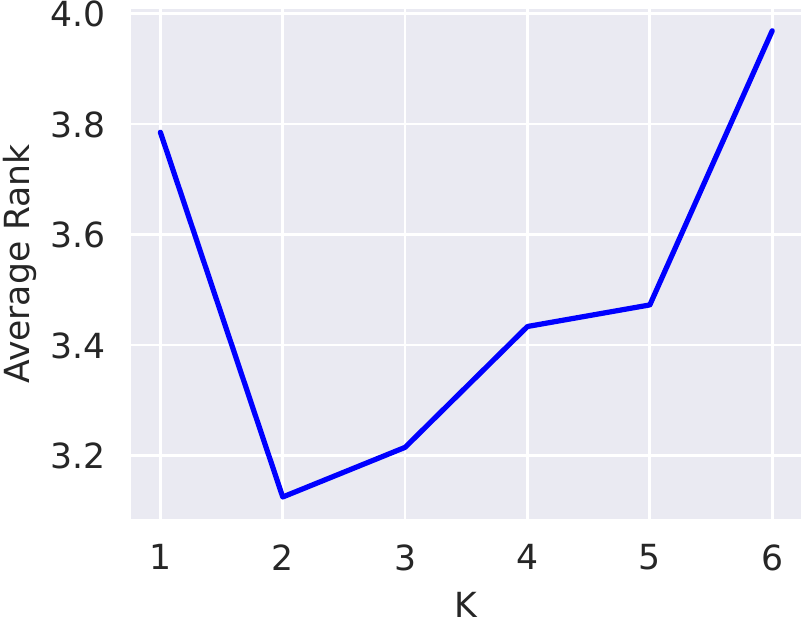}
    \caption{20 labels per class}
    \vspace{-10pt}
  \end{subfigure}%
  \caption{Impact of K.}
  \vspace{-10pt}
  \label{fig:impactk}
\end{figure}

\begin{table}[t]
    \centering
    \scriptsize
    \caption{Comparison of negative and random sampling on the full UCR datasets (top). Comparison with the variant without batch training on 121 UCR datasets (bottom), on which w/o batch training does not suffer from GPU memory explosion. $\blacktriangle$ suggests negative sampling is significantly better than the other algorithms.}
    \label{tab:ablation}
    \vspace{-10pt}
    \setlength{\tabcolsep}{5.0pt}
    \begin{tabular}{l|llll}
    \toprule
     
 \diagbox [width=12.5em,trim=l] {Algorithm}{Labels}& 10 & 20 & 30 & 40 \\
    \midrule
    \midrule
     random sampling & 1.637$\blacktriangle$ & 1.609$\blacktriangle$ & 1.547 & 1.582$\blacktriangle$ \\
     negative sampling & \textbf{1.363} & \textbf{1.391} & \textbf{1.453} & \textbf{1.418} \\

     \midrule
      random sampling & 2.152$\blacktriangle$ & 1.979$\blacktriangle$ & 1.868 & 1.926$\blacktriangle$ \\
      w/o batch training & 2.068$\blacktriangle$ & 2.277$\blacktriangle$ & 2.331$\blacktriangle$ & 2.322$\blacktriangle$ \\
      negative sampling & \textbf{1.780} & \textbf{1.744} & \textbf{1.802} & \textbf{1.752} \\

     \bottomrule
    \end{tabular}
    \vspace{-20pt}
\end{table}

\vspace{-5pt}
\subsection{Analysis of Graph Structure}
\label{sec:exp4}
For \textbf{RQ4}, we vary the hyperparameters of graph construction. In Figure~\ref{fig:impactalpha}, we vary $\alpha$ from the set $\{0.1, 0.3, 0.5, 0.7, 0.9, 1.1\}$ with $K$ fixed to be $3$. We observe a significant performance drop when $\alpha=0.1$. This is because when $\alpha \to 0$, each node will aggregate more information from the neighbors, and the resulting representation will become less distinguishable. There is also a performance drop when $\alpha=1.1$. A larger $\alpha$ will reduce the similarity information in the model, which suggests that incorporating similarity information is indeed helpful. In Figure~\ref{fig:impactk}, we vary $K$ from the set $\{1, 2, 3, 4, 5, 6\}$ with $\alpha$ fixed to be $0.3$. The best performance is achieved with $K=2$. When $K=1$, our framework reduces to backbone itself since there will be only self-connections in the graph. When $K$ becomes larger, the performance also drops since aggregating more information from the neighbors makes the representations less distinguishable. Overall, we find that incorporating an appropriate amount of similarity information lead to the best performance. 

\vspace{-5pt}
\subsection{Ablation Study}
\label{sec:exp5}
For \textbf{RQ5}, we consider two ablations to study the effectiveness of negative sampling. First, we consider a variant that uses random sampling for batch training. Second, we consider a variant that puts all the time-series into GPU memory without batch training. Unfortunately, this variant will explode the GPU memory for 7 of the datasets so that only 121 datasets are reported. We perform a grid-search of $K$ and find that $K=10$ achieves the best performance.

Table~\ref{tab:ablation} summarizes the results. First, we observe that negative sampling outperforms random sampling significantly. Second, the variant without batch training does not perform well. A possible explanation is that the graph is pre-constructed so that each time-series can only interact with a limited number of connected neighbors, which may lead to sub-optimal performance.

\begin{table*}[t]
    \centering
    \scriptsize
    \caption{Classification accuracy on multivariate time-series datasets across different numbers of training labels per class. For KickvsPunch, we only report the performance up to 15 labels since there are less than 15 labels per class. TapNet is not reported on Character Trajectories and Netflow due to memory explosion.}
    \label{tab:performancemulti}
    \vspace{-10pt}
    \setlength{\tabcolsep}{-0.05pt}
    \begin{tabular}{l|l|ccccccccccc}
    \toprule
     
 {Dataset} & \diagbox [width=7em,trim=l] {Algorithm}{Labels} & 5 & 10 & 15 & 20 & 25 & 30 & 35 & 40 & 45 & 50 \\
    \midrule
    \midrule
     \multirow{7}{*}{\shortstack{Character\\ Trajectories}} & DTW & .847$\pm$.014 & .881$\pm$.005 & .895$\pm$.009 & .900$\pm$.014 & .908$\pm$.014 & .907$\pm$.013 & .906$\pm$.010 & .906$\pm$.007 & .909$\pm$.010 & .913$\pm$.008 \\
      & ResNet & .834$\pm$.024 & .898$\pm$.017 & .920$\pm$.010 & .937$\pm$.010 & .939$\pm$.008 & .941$\pm$.009 & .949$\pm$.008 & .956$\pm$.011 & .958$\pm$.007 & .958$\pm$.007 \\
      & InceptionTime & .883$\pm$.010 & .939$\pm$.007 & .947$\pm$.006 & \textbf{.968$\pm$.006} & .964$\pm$.003 & .974$\pm$.005 & .979$\pm$.003 & .978$\pm$.005 & .979$\pm$.001 & \textbf{.986$\pm$.003} \\
      & TapNet & - & - & - & - & - & - & - & - & - & - \\
      & SimTSC-S & .894$\pm$.020 & .939$\pm$.009 & .949$\pm$.007 & .947$\pm$.017 & .964$\pm$.011 & .975$\pm$.003 & .977$\pm$.011 & .975$\pm$.004 & .981$\pm$.007 & .982$\pm$.005 \\
      & SimTSC-I & \textbf{.914$\pm$.012} & .944$\pm$.009 & .951$\pm$.015 & .953$\pm$.012 & \textbf{.969$\pm$.011} & \textbf{.978$\pm$.006} & \textbf{.981$\pm$.007} & .979$\pm$.005 & .977$\pm$.008 & .980$\pm$.003 \\
      & SimTSC-T & .903$\pm$.014 & \textbf{.946$\pm$.005} & \textbf{.957$\pm$.011} & .964$\pm$.009 & .967$\pm$.012 & .973$\pm$.009 & .976$\pm$.009 & \textbf{.981$\pm$.006} & \textbf{.983$\pm$.008} & \textbf{.986$\pm$.004} \\
      
      \midrule
      
      \multirow{7}{*}{ECG} & DTW & .605$\pm$.124 & .670$\pm$.086 & .740$\pm$.112 & .755$\pm$.103 & .805$\pm$.043 & .825$\pm$.050 & .805$\pm$.053 & .800$\pm$.057 & .805$\pm$.053 & .800$\pm$.057 \\
      & ResNet  & .745$\pm$.048 & .795$\pm$.037 & .805$\pm$.058 & .800$\pm$.079 & \textbf{.860$\pm$.030} &  \textbf{.855$\pm$.048} & .850$\pm$.052 & \textbf{.855$\pm$.029} & .830$\pm$.037 & \textbf{.870$\pm$.029} \\
      & InceptionTime & .750$\pm$.045 & .805$\pm$.033 & .785$\pm$.020 & .800$\pm$.037 & .820$\pm$.037 &  .830$\pm$.043 & .825$\pm$.016 & .850$\pm$.027 & \textbf{.855$\pm$.0010} & .850$\pm$.016 \\
      & TapNet & .770$\pm$.043 & .780$\pm$.012 & .755$\pm$.025 & .795$\pm$.048 & .810$\pm$.037 &  .795$\pm$.029 & .785$\pm$.025 & .815$\pm$.037 & .830$\pm$.019 & .845$\pm$.024 \\
      & SimTSC-S & .795$\pm$.043 & .810$\pm$.020 & \textbf{.855$\pm$.040} & \textbf{.840$\pm$.051} & .830$\pm$.056 & .840$\pm$.020 & \textbf{.860$\pm$.041} & .825$\pm$.047 & .830$\pm$.071 & .860$\pm$.025 \\
      & SimTSC-I & .790$\pm$.062 & .765$\pm$.072 & .830$\pm$.070 & .730$\pm$.159 & .740$\pm$.087 & .800$\pm$.091 & .830$\pm$.048 & .750$\pm$.052 & .790$\pm$.108 & .735$\pm$.108 \\
      & SimTSC-T & \textbf{.810$\pm$.041} & \textbf{.815$\pm$.046} & .770$\pm$.108 & .815$\pm$.115 & .730$\pm$.118 & .745$\pm$.075 & .745$\pm$.099 & .780$\pm$.051 & .775$\pm$.071 & .710$\pm$.101 \\
      
      \midrule
      
      \multirow{7}{*}{KickvsPunch} & DTW & .433$\pm$.082 & .433$\pm$.082 & .433$\pm$.082 & - & - & - & - & - & - & - \\
      & ResNet & .667$\pm$.183 & \textbf{.833$\pm$.149} & .833$\pm$.183 & - & - & - & - & - & - & - \\
      & InceptionTime & .667$\pm$.000 & .533$\pm$.125 & .567$\pm$.226 & - & - & - & - & - & - & - \\
      & TapNet & .700$\pm$.125 & .767$\pm$.082 & .733$\pm$.013 & - & - & - & - & - & - & - \\
      & SimTSC-S & \textbf{.733$\pm$.200} & .767$\pm$.133 & \textbf{.867$\pm$.125} & - & - & - & - & - & - & - \\
      & SimTSC-I & .700$\pm$.125 & \textbf{.833$\pm$.105} & .800$\pm$.125 & - & - & - & - & - & - & - \\
      & SimTSC-T & .600$\pm$.133 & .767$\pm$.133 & .767$\pm$.082 & - & - & - & - & - & - & - \\
      
      \midrule
      
      \multirow{7}{*}{NetFlow} & DTW & .611$\pm$.016 & .559$\pm$.128 & .607$\pm$.132 & .595$\pm$.118 & .546$\pm$.103 & .568$\pm$.125 & .523$\pm$.154 & .481$\pm$.203 & .503$\pm$.217 & .504$\pm$.214 \\
      & ResNet & .613$\pm$.074 & .714$\pm$.063 & .749$\pm$.022 & .763$\pm$.038 & .739$\pm$.058 & .767$\pm$.050 & .769$\pm$.054 & .767$\pm$.049 & .787$\pm$.026 & .797$\pm$.039 \\
      & InceptionTime & .418$\pm$.052 & .456$\pm$.046 & .484$\pm$.058 & .618$\pm$.049 & .642$\pm$.036 & .657$\pm$.024 & .678$\pm$.014 & .675$\pm$.036 & .681$\pm$.018 & .681$\pm$.015 \\
      & TapNet & - & - & - & - & - & - & - & - & - & - \\
      & SimTSC-S & .519$\pm$.108 & .720$\pm$.071 & .705$\pm$.055 & .709$\pm$.089 & .738$\pm$.082 & .786$\pm$.036 & .765$\pm$.091 & .790$\pm$.045 & .784$\pm$.063 & .799$\pm$.047 \\
      & SimTSC-I & .766$\pm$.043 & .788$\pm$.036 & .689$\pm$.139 & \textbf{.776$\pm$.042} & .731$\pm$.084 & .755$\pm$.104 & \textbf{.834$\pm$.037} & .798$\pm$.066 & .810$\pm$.065 & .839$\pm$.035 \\
      & SimTSC-T & \textbf{.769$\pm$.052} & \textbf{.805$\pm$.035} & \textbf{.785$\pm$.101} & .766$\pm$.095 & .\textbf{745$\pm$.092} & \textbf{.825$\pm$.029} & .801$\pm$.065 & \textbf{.827$\pm$.059} & \textbf{.847$\pm$.023} & \textbf{.852$\pm$.028} \\
      
      
      
     \bottomrule
    \end{tabular}
    \vspace{-10pt}
\end{table*}

\subsection{Results on Multivariate Datasets}
\label{sec:exp6}
For \textbf{RQ6}, we evaluate SimTSC on multivariate time-series classification tasks in Table~\ref{tab:performancemulti}. First, SimTSC outperforms ResNet and DTW with very few labels. Given 5, 10, or 15 labels, at least one of the SimTSC variants achieves the best performance. Second, when more labels are given, the similarity information helps on some datasets but worsens some others' performance. SimTSC delivers poor performance on ECG in the transductive setting given 50 labels, while we observe a consistent improvement on CharacterTrajectories and NetFlow.

\subsection{Visualization of Learned Representations}
\label{sec:exp7}
\begin{figure}[t]
  \centering
  \begin{subfigure}[b]{0.240\textwidth}
    \centering
    \includegraphics[width=0.99\textwidth]{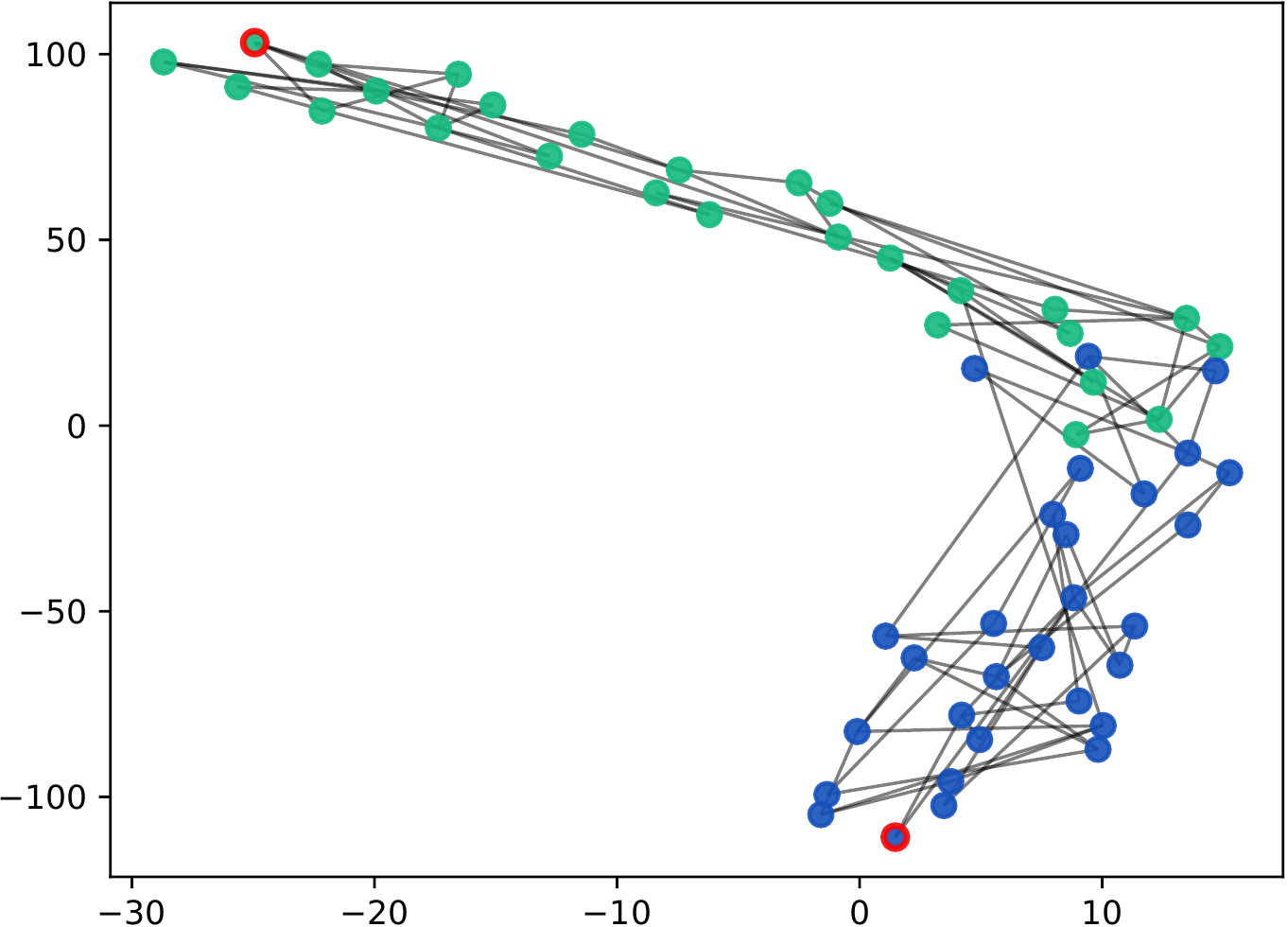}
    \caption{ResNet (83\% accuracy)}
  \end{subfigure}%
  \begin{subfigure}[b]{0.240\textwidth}
    \centering
    \includegraphics[width=0.99\textwidth]{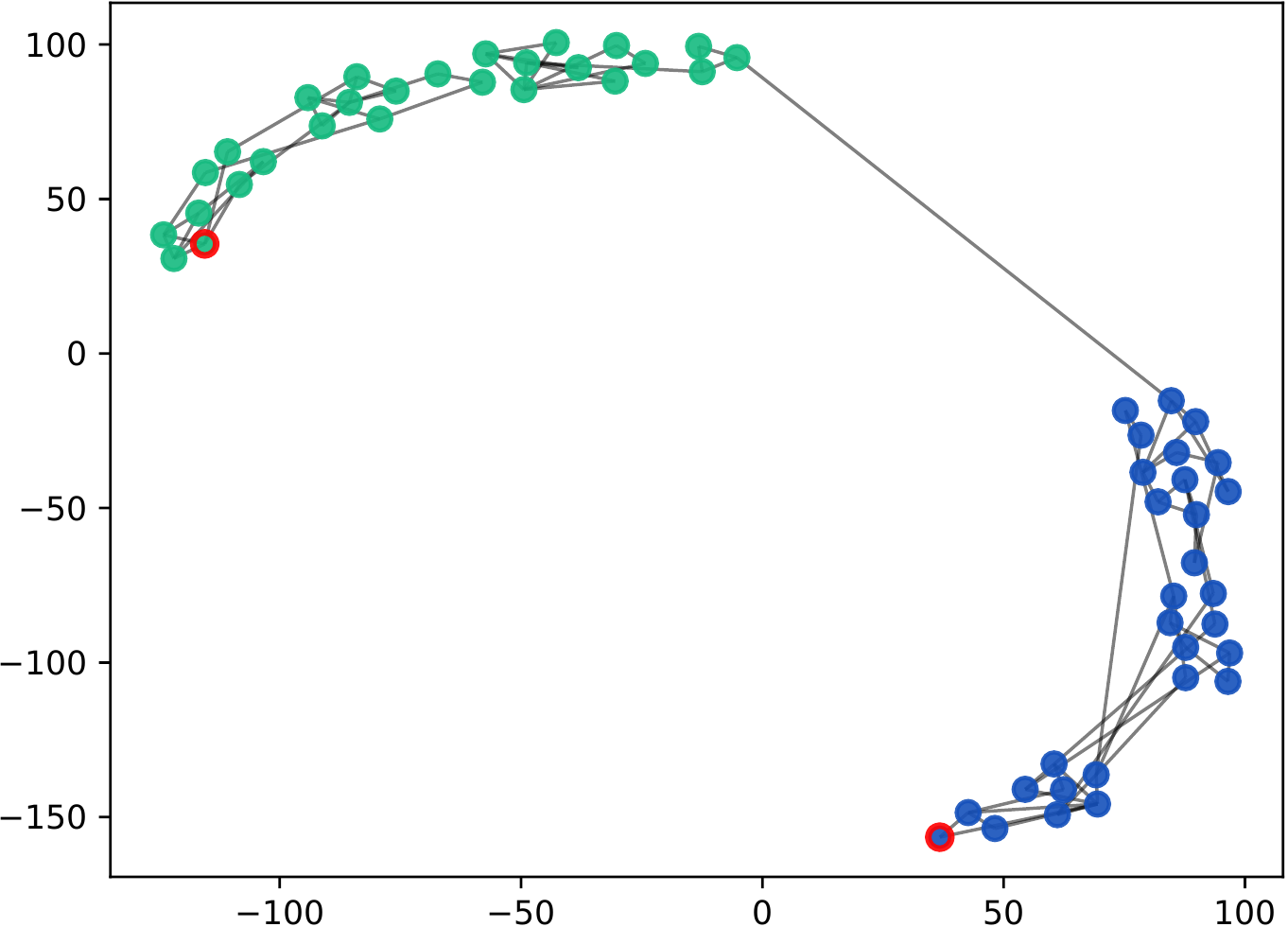}
    \caption{SimTSC (100\% accuracy)}
  \end{subfigure}%
  \vspace{-10pt}
  \caption{Learned representations of ResNet and SimTSC on Coffee with 56 time-series, two classes marked in blue and green, respectively, and only one time-series labeled in each class (circled in red).}
  \label{fig:visual}
  \vspace{-20pt}
\end{figure}

To answer \textbf{RQ7}, we conduct a case study on the Coffee dataset from UCR Archive. Figure~\ref{fig:visual} visualizes the learned representations of SimTSC and ResNet as well as the constructed graphs. We observe that there is an overlap between the two classes' representations learned by ResNet. Thus, ResNet cannot distinguish those overlapped time-series and only gives 83\% accuracy. Whereas, the representations learned by SimTSC form clear clusters so that SimTSC achieves 100\% accuracy. A possible reason is that the overlapped time-series tend to be close to those in the same class in terms of DTW, and thus their representations are corrected by aggregation. SimTSC achieves better accuracy by jointly performing feature extraction and aggregation.

\section{Related Work}
\textbf{Deep Learning for TSC.} 
Prior deep learning models for TSC can be mainly grouped into (1) learning representations of time-series in an unsupervised manner and then applying a classifier to the learned representations~\cite{mittelman2015time}, and (2) training a classifier in an end-to-end fashion~\cite{geng2018cost,bagnall2017great,karim2017lstm}. However, the existing models mainly focus on feature extraction but cannot capture similarity information. Although~\cite{wu2020connecting} has introduced GNNs to capture time-series dependencies, they focus on time-series forecasting and can only model the dependencies among multivariate time-series. Whereas, we adopt GNNs to jointly perform feature extraction and model the time-series similarities for TSC.


\textbf{Similarity-Based TSC Methods.}
Similarity-based methods first obtain time-series similarities and then use a $k$-NN classifier for classification~\cite{nicolae2016similarity,wu2018random}. Recently, some similarity learning approaches have been proposed to learn the similarities~\cite{abid2018learning,grabocka2018neuralwarp}. However, they often rely on a separate procedure for classification. In contrast, we connect time-series similarity with deep models under a unified framework.


\textbf{Leveraging Unlabeled Data in Time-Series.} Prior work has explored semi-supervised learning~\cite{wang2019time}, domain adaptation~\cite{li2021learning}, and anomaly detection~\cite{li2020pyodds,zhao2019pyod,lai2021tods,lai2021revisiting} on the unlabeled data, most of which do not target deep models. Recently, \cite{jawed2020self} proposes a deep TSC method with auxiliary forecasting tasks. \cite{zhang2020tapnet} augments networks with task-adaptive projection. However, they do not support batch training. Unlike the previous work, our framework brings benefits to not only semi-supervised setting but also the supervised setting.

\section{Conclusions and Future Work}

This work explores connecting the research efforts of time-series similarity measuring and deep learning for TSC. To jointly model feature extraction and similarity information, we formulate TSC as a node classification problem in graphs and introduce GNNs on the top of a backbone to enable end-to-end training. We instantiate our framework with ResNet and DTW with extensive experiments on the full 128 UCR datasets and several multivariate datasets. Experimental results suggest that incorporating similarity information can improve deep models significantly. In the future, we will investigate differentiable DTW~\cite{cuturi2017soft} for graph construction.

\section*{Acknowledgements}
The work is, in part, supported by NSF (\#IIS-1849085, \#IIS-1900990, \#IIS-1750074). The views and conclusions in this paper are those of the authors and should not be interpreted as representing any funding agencies.

\bibliographystyle{siam}
\bibliography{ref}


\newpage
\appendix

\begin{table}[t]
    \centering
    \scriptsize
    \caption{Statistics of the multivariate time series datasets.}
    \label{tab:stats}
    \vspace{-10pt}
    \setlength{\tabcolsep}{0pt}
    \begin{tabular}{l|ccccc}
    \toprule
     
 \diagbox [width=10 em,trim=l] {Dataset}{Attribute} & Data Size & Dimension & \# Classes & Length \\
 
    \midrule
    \midrule
     Character Trajectories & 2858 & 3 & 20 & 109-205 \\
     ECG & 200 & 2 & 2 & 39-152 \\
     KickvsPunch & 26 & 62 & 2 & 274-841 \\
     NetFlow & 1337 & 2 & 4 & 50-997 \\
   
     \bottomrule
    \end{tabular}
    \vspace{-15pt}
\end{table}

\section{Experimental Details}
\label{sec:appendix}

\subsection{Datasets}
We use both univariate datasets and multivariate datasets in our experiments. We provide detailed descriptions below.

\textbf{Univariate Time-Series Datasets.} The experiments are conducted on the full 128 datasets from the UCR Time Series Classification Archive\footnote{\url{https://www.cs.ucr.edu/~eamonn/time_series_data_2018/}}~\cite{dau2019ucr}. These datasets are collected from various domains, including Image, Spectro, Sensor, Simulated, Device, Motion, ECG, Traffic, EOG, HRM, Trajectory, and Hemodynamics. These datasets also have diverse data sizes, numbers of classes, and lengths. For example, ElectricDevices, one of the largest datasets in the archive, has 16,637 time-series in total, while BeetleFly only has 40 time-series. Similarly, ShapesAll has 60 classes, while many other datasets have only 2 or 3 classes. The time-series length can also be up to 2,709 and can also be as short as 60. Moreover, 11 of the datasets have varying lengths for different time-series. Thus, the UCR datasets provide a rigorous test for time-series classification.

\textbf{Multivariate Time-Series Datasets.} We focus on four multivariate datasets from different domains with varying characteristics.

\begin{itemize}
    \item \textbf{Character Trajectories:} is a handwriting dataset captured at 200Hz by a WACOM tablet. The three dimensions are the positions of the x-axis, the y-axis, and the pen tip force. The data has been numerically differentiated, and Gaussian smoothed, with a sigma value of 2.
    \item \textbf{ECG:} traces the electrical activity recorded during heartbeats. The two classes are the normal heartbeat and the myocardial infarction.
    \item \textbf{KickVsPunch:} is a motion capture dataset collected by CMU-MOCAP. Each dimension is a motion marker. The two classes are the two actions, i.e., the kick and the punch.
    \item \textbf{NetFlow:} is the traffic flow of websites. 
\end{itemize}
We summarize the statistics of the above datasets in Table~\ref{tab:mlcfcn}.

\subsection{Data Preprocessing}

For both univariate and multivariate time-series data, we store them as 3-dimensional Numpy arrays, where the first dimension is the number of time-series, the second dimension is the number of time-series (for univariate time-series, there is only one time-series), and the third dimension is the length of the time-series. We impute missing values with zeros for the datasets with varying lengths to make the time-series have the same length. All the time-series are z-normalized before feeding into the models.

\subsection{Data Splitting}
The original splits in UCR datasets are very diverse. For example, ElectricDevices has 8,926 training samples, while ECGFiveDays only has 23 training samples. This makes it difficult to understand how the models will behave under different amounts of supervision. Specifically, when an algorithm performs well, it is hard to tell how it performs when we have very few labels and how it will perform when we have enough labels. Motivated by few-shot learning, we simulate the settings where different numbers of labels per class are given. Specifically, we merge the original training and testing data to create new splits as follows. First, we merge the Numpy arrays of training and testing data, where the training data is put before the testing data. Second, we randomly shuffle all the indices. Third, we separate out the last 20\% of the shuffled indices as the hold-out testing time-series. Fourth, given the target number of labels per class, we iterate over the first 80\% of the shuffled indices sequentially until we find enough number of labeled data for each class. It is possible that we can not find enough labeled data for some classes even after iterating all the indices. In this case, we simply use as many labels as we can. For example, for a class A and a target number 10, if the number of class A data in the first 80\% of the shuffled indices is only eight, we simply use eight training data for class A. The remaining data in the first 80\% of the shuffled indices will serve as the third split (unlabeled data).

The above three splits are used to simulate the supervised, inductive semi-supervised, and transductive semi-supervised settings defined in Section~\ref{sec:21}. The performance will be evaluated on the hold-out testing set. The above splitting procedure is applied to all the univariate and multivariate time-series data. Since data splitting may significantly affect the performance, particularly when we have very few labels, we run each experiment 5 times on different splits.

\subsection{Dynamic Time Warping}
Dynamic Time Warping (DTW) is s standard algorithm for measuring the similarity between two time-series. The main idea of DTW is to calculate the optimal match between two time-series such that the sum of matched series has the smallest values. In this work, the DTW is computed based on a Python wrapper of The UCR Suite\footnote{\url{https://www.cs.ucr.edu/~eamonn/UCRsuite.html}}. This suite provides a highly efficient C++ implementation of DTW via dynamic programming. For all the datasets, we set the size of the warping window to be 100. If either of the two time-series is shorter than 100, we use the shortest time-series length.

For multivariate datasets, we compute independent DTW. Specifically, we first compute the DTW for each pair of univariate time-series and then sum them up to represent the distance between two multivariate time-series. We have uploaded the pre-computed DTW to Google Drive for reproducibility.

Since most of the datasets are small, it will not take much time to compute the full similarity matrix. For most of the datasets, the computation of DTW can be finished in minutes. For some larger datasets, it takes at most a day using one CPU core.

\subsection{Neural Architecture of Backbone}
\begin{figure}[H]
     \vspace{-12pt}
  \centering
    \includegraphics[width=0.22\textwidth]{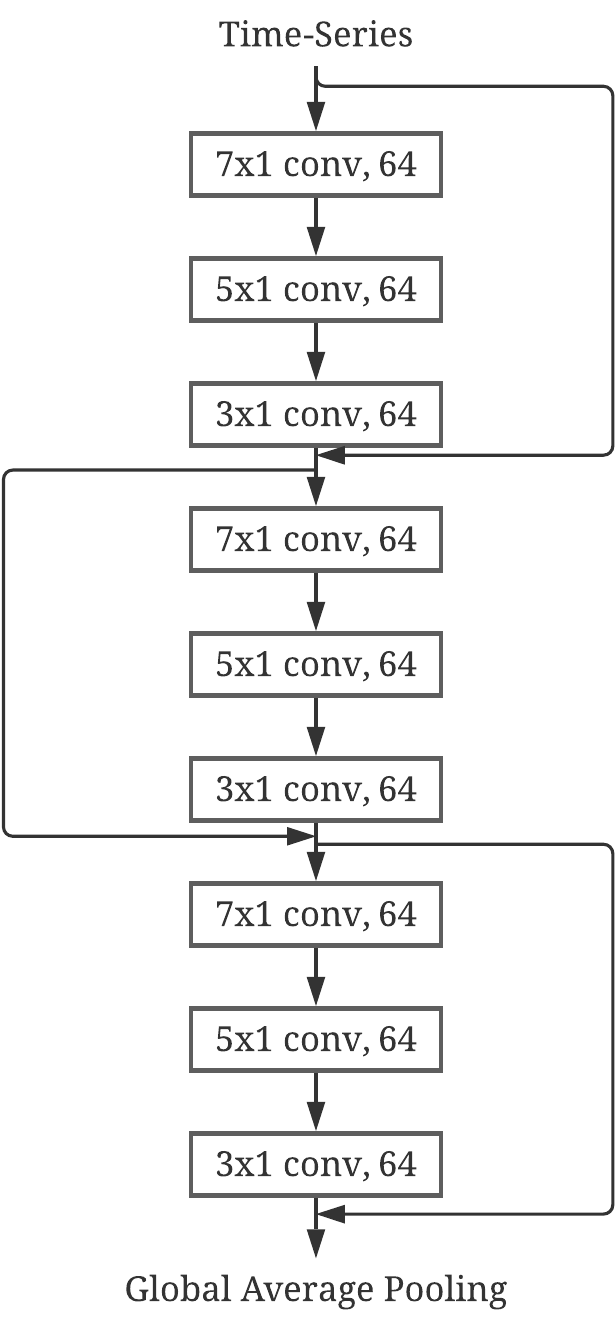}
  \vspace{-12pt}
  \caption{Neural architecture of ResNet. Each layer is followed by a batch normalization layer and a ReLU activation.}
  \label{fig:resnet}
   \vspace{-12pt}
\end{figure}

We use PyTorch to implement all the neural networks. For most of the experiments, we use ResNet as the backbone. The network consists of three residual blocks. Each residual block consists of three 1-D convolution layers. The kernel sizes of the three convolution layers are 7, 5, and 3, respectively. After each convolution layer, we use a 1-D batch normalization layer to stabilize training, followed by a ReLU activation function. The number of channels is set to be 64 for all the convolution layer. We find that using more channels will lead to unsatisfactory performance with very little training data due to overfitting issue. A skip connection is added in each block to enable direct flow to alleviate the gradient vanishing issue. The three blocks are stacked sequentially to perform feature extraction. To reduce the feature dimension, we add a global average pooling layer to the last residual block's output. For our SimTSC, we directly use the global average pooling layer's output as the extracted features. For the ResNet baseline, these features will be further processed by a fully-connected layer with a softmax activation for classification purposes. The weights of the networks are initialized with the default initializers in PyTorch. Figure~\ref{fig:resnet} summarizes the neural architecture of ResNet.

For the MLP baseline, we use four fully-connected layers with a hidden size of 500. We use a ReLU activation function after each fully-connected layer followed by a dropout layer to avoid overfitting. The dropout rates are set to be 0.1, 0.2, 0.2, 0.3 for the four layers. We flatten the time-series into one dimension so that it can be directly used in MLP. Different from convolution layers, MLP can not capture temporal information, which leads to unsatisfactory performance. For the FCN baseline, we use three 1-D convolution layers. Each convolution layer is followed by a batch normalization layer. The kernel sizes are set to be 7, 5, and 3. Similar to ResNet, the number of channels is set to 64. For both MLP and PyTorch, we use the default initializers in PyTorch to initialize the weights.

Note that, for all the experiments, we use exactly the same architecture for the backbone of SimTSC and the baseline. The only difference of SimTSC is adding a GCN layer on the top of the backbone. Thus, the comparison is fair. The performance gain is solely attributed to modeling the similarity information.

\subsection{Graph Convolution Layers}
The graph convolution layers are implemented based on the standard implementation of GCN\footnote{\url{https://github.com/tkipf/pygcn}}. Specifically, each GCN layer takes node features and an adjacency matrix as the input, where the adjacency matrix is a sparse tensor. Then it aggregates the neighbors' features by performing matrix multiplication of the input features, weights, and the adjacency matrix. Finally, a bias term is added to the obtained features. For multiple GCN layers, we add a dropout layer after each GCN layer to avoid overfitting.

The adjacency matrix used in the graph convolution layers is constructed as follows. We maintain a full pre-computed similarity matrix in the memory. In each update step, we sample a batch of indices for training. We then use the sampled index to obtain a submatrix of the full similarity matrix. This submatrix will only contain the indices in this batch. Further, we rank each row's values in ascending order and only keep the top-$K$ similar neighbors for each row. We finally use the top-$K$ neighbors to construct the graph, which is represented by a sparse matrix. The above constructing procedure is efficient since we only need to take care of a batch of indices instead of all the indices.

\subsection{Hyperparameter Settings}
We summarize the hyperparmeters of graph, optimizer, and how we train SimTSC and all the baselines as follows.
\begin{itemize}
    \item \textbf{Graph Construction:} We set the scaling factor $\alpha=0.3$ and number neighbors for each node $K=3$.
    \item \textbf{Graph Convolution:} We use one GCN layer for most of the experiments. For multiple GCN layers, the feature dimension is set to be 64, and the dropout rate is set to be 0.5. 
    \item \textbf{Optimizer:} We use Adam optimizer. The learning rate is set to be 0.0001. The $\epsilon$ is set to be $10^{-8}$.
    \item \textbf{Training Procedure:} For SimTSC and all the baselines, we use the model that achieves the best performance on the training data for evaluation. Specifically, we calculate the accuracy based on the training data after each training epoch and store the model's weights with the highest accuracy. Then the stored weights will be reloaded for evaluation purposes. For all the models, we train 500 epochs. While validating on the training data may lead to overfitting, we find in practice that it works better than separating a validation set from the training data. This is because a separated validation set will be too small to perform a meaningful evaluation.
\end{itemize}

\subsection{Hardware and Software Descriptions}
We conduct all the experiments on a server with two AMD EPYC 7282 16-Core processors, four GeForce RTX 3090 GPUs, and 252 GB memory. We use Ubuntu 18.04.5 LTS system and PyTorch 1.7.0.

\end{document}